\documentclass[pdflatex,sn-mathphys-num]{sn-jnl}

\usepackage{graphicx}%
\usepackage{multirow}%
\usepackage{amsmath,amssymb,amsfonts}%
\usepackage{amsthm}%
\usepackage{mathrsfs}%
\usepackage[title]{appendix}%
\usepackage{xcolor}%
\usepackage{textcomp}%
\usepackage{manyfoot}%
\usepackage{booktabs}%
\usepackage{algorithm}%
\usepackage{algorithmicx}%
\usepackage{algpseudocode}%
\usepackage{listings}%
\usepackage{color}
\usepackage{array}
\usepackage{caption}
\usepackage{subfigure, epsfig, hyperref}
\usepackage{amsthm}
\usepackage{tabularx}
\usepackage{enumitem}
\usepackage{xcolor}
\usepackage{subcaption}
\usepackage{pifont}
\usepackage{makecell}
\usepackage{bbm}
\usepackage{diagbox}
\usepackage{multirow}
\usepackage{array}
\usepackage{amsmath}
\usepackage{dsfont}
\usepackage[normalem]{ulem}
\useunder{\uline}{\ul}{}
\usepackage{placeins}
\usepackage{colortbl}
\definecolor{mybackground}{RGB}{247,237,247}
\usepackage{etoolbox}
\usepackage{geometry}
\usepackage[most]{tcolorbox}
\makeatletter
\patchcmd{\@author}{\large}{\small}{}{}
\makeatother

\theoremstyle{thmstyleone}%

\theoremstyle{thmstyletwo}%

\theoremstyle{thmstylethree}%

\begin{document}

\title{\Large Cost-effective Instruction Learning for Pathology Vision and Language Analysis}

\author[1,2]{\small Kaitao Chen}
\equalcont{\small These authors contributed equally to this work.}

\author[1]{\small Mianxin Liu}
\equalcont{\small These authors contributed equally to this work.}

\author[1]{\small Fang Yan}
\author[3]{\small Lei Ma}
\author[1]{\small Xiaoming Shi}
\author[1]{\small \fnm{Lilong} \sur{Wang}}
\author[1]{\small Xiaosong Wang}
\author[4]{\small Lifeng Zhu}
\author[5]{\small Zhe Wang}
\author[6]{\small \fnm{Mu} \sur{Zhou}}
\author*[1,7]{\small Shaoting Zhang}\email{zhangshaoting@pjlab.org.cn}

\affil[1]{\small \orgaddress{Shanghai Artificial Intelligence Laboratory, Shanghai, China}}
\affil[2]{\small \orgaddress{School of Computer Science, Fudan University, Shanghai, China}}
\affil[3]{\small \orgaddress{National Biomedical Imaging Center, College of Future Technology, Peking University, Beijing, China}}
\affil[4]{\small \orgaddress{Ruijin Hospital, Shanghai Jiaotong University School of Medicine, Shanghai, China}}
\affil[5]{\small \orgaddress{Department of Pathology, State Key Laboratory of Cancer Biology, Xijing Hospital, Xi'an, China}}
\affil[6]{\small \orgaddress{Department of Computer Science, Rutgers University, New Jersy, US}}
\affil[7]{\small \orgaddress{Centre of Perceptual and Interactive Intelligence under the InnoHK, Hong Kong SAR, China}}

\abstract{The advent of vision-language models fosters the interactive conversations between AI-enabled models and humans. Yet applying these models into clinics must deal with challenges around large-scale training data, financial, and computational resources. Here we propose CLOVER, a cost-effective instruction learning framework for conversational pathology. CLOVER trains a lightweight module and uses instruction tuning while freezing the parameters of the large language model. Instead of using costly GPT-4, we propose well-designed prompts on GPT-3.5 for building generation-based instructions, emphasizing the utility of pathological knowledge derived from the Internet source. We construct a high-quality set of template-based instructions in the context of digital pathology. From two benchmark datasets, our findings reveal the strength of hybrid-form instructions in pathological visual question-answer. CLOVER outperforms baselines that possess 37 times more training parameters and exhibits few-shot capacity in the external clinical dataset. CLOVER could thus accelerate the adoption of rapid conversational applications in digital pathology.}

\maketitle

\section{Introduction}

The rise of vision-language model (VLM) opens remarkable opportunities to analyze pathological images in a visual question-answering manner \cite{zhang2024data, prefixT, M2M}. This profound progress of multi-modal data integration leverages the power of large language model (LLM) on cognition, reasoning, and content generation \cite{llm1, llm2, CoT}. In essence, LLMs are large-scale parametric networks trained on vast amounts of data, enabling them to generate human-like responses and achieve remarkable accuracy. ChatGPT \cite{chatgpt} and GPT-4 \cite{gpt4} are examples to simulate the conversational interaction between AI-enabled models and humans. The generated conversational records can be re-used to guide the visual-language model refinement \cite{llava, Flamingo, FROMAGe, mplug-owl}, opening avenues for downstream tasks across domains \cite{zhang2024data}. In the landscape of digital pathology, providing in-depth language descriptions of cell morphology, tissue status, and treatment suggestions, equipped with human-like interactions, could enhance tissue understanding, characterization, and decision making in various clinical scenarios \cite{song2023artificial, artificial, digital, wang2024editorial, zhang2024challenges}.

Emerging pathological vision-language models (PVLMs) (for instance, LLaVA-Med \cite{llava-med} and Quilt-LLaVA \cite{Quilt-LLaVA}) have demonstrated their utility in analyzing medical imaging. However, it is widely known that building a capable PVLM demands an excessive training data, human labour, financial, and computational resources \cite{Pmc-llama, Med-flamingo, musk, lu2024visual, lu2024multimodal, xu2024multimodal}. Extending general-purpose models into pathology-oriented model starts with using pathological vision-language datasets. These datasets consist of (i) image-text pairs (image-caption) dataset and (ii) instruction (image-question-answering) dataset \cite{pmc, Quilt-1M, Twitter, lu2024visual}. The image-text dataset aligns visual and language features \cite{clip}, providing rich semantic information for pathological image contents \cite{gao2024aligning, textguided, pathology-geo}. Meanwhile, instruction dataset is crucial for activating LLMs to complete the visual-language question answering. Yet the process of instruction generation incurs a considerable financial cost by using GPT-4 (nearly \$9,000) \cite{Quilt-LLaVA}. In addition to the instruction data, model optimization also requires a substantial compute support. Directly tuning a billion-parameter LLM demands high computational resources and is impractical to achieve on consumer-grade GPUs. To overcome these bottlenecks, fundamental questions are urged to be addressed towards building an effective but low-cost PVLM: (i) How can we come up with a lightweight tuning approach to complement the LLaVA-like tuning on large training parameters? (ii) How can we generate a useful instruction dataset in a cost-effective way? (iii) How can we achieve few-shot generalized learning ability of PVLM for clinical applications without using a large-scale instruction dataset?

In this study, to address these questions, we propose a cost-effective learning framework for accurate pathology vision and language inference named as CLOVER. Our study has made multifaceted contributions: \textbf{First}, we propose an effective PVLM training framework particularly at low computational resources and financial spending. Different from the LLaVA-like tuning (for instance, LLaVA-Med), we emphasize the use of BLIP-2 \cite{blip2} to serve as an alternative choice for the cost-effective lightweight inference. We find that tuning the wide spectrum of LLM's parameters is unnecessary for building a sufficiently usable PVLM. \textbf{Second}, the value of our designed PVLM-oriented prompt is pronounced on GPT-3.5 instead of the advanced GPT-4. Low-cost CLOVER model outperforms those models trained with LLM tuning on the instruction data generated by GPT-4 in multiple settings. We also recognize template-based instructions without relying on GPT can supplement generation-based instructions. A combined use of these two type instructions can further enhance CLOVER's understanding capabilities. \textbf{Third}, we confirm that CLOVER can be effectively developed with a limited scale of instructions towards a frugal and real-world application. This is evidenced by the appealing performance in fine-tuning and zero-shot experiments on two visual question answering datasets, as well as in the few-shot experiment on a real-world clinical dataset.

\begin{figure}[!t]
    \centering
    \includegraphics[width=\linewidth]{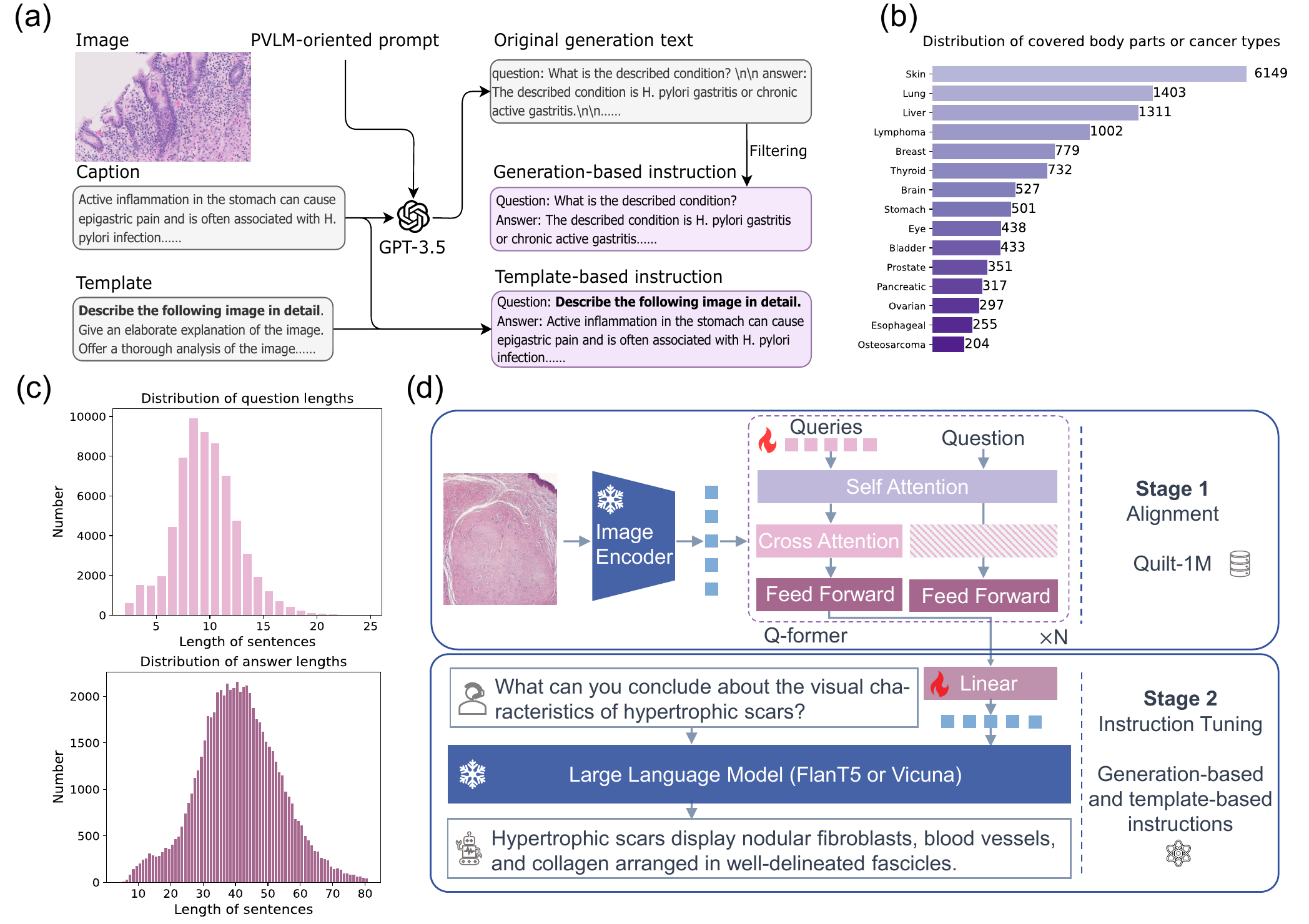}
    \caption{(a) The workflow of instruction generation. We propose a low-cost solution of instruction data generation carefully designed for analyzing pathological data. Generation-based instructions are created by rewriting image captions into a question-and-answer format using low-cost GPT-3.5. In template-based instructions, the question is a descriptive prompt, while the answer is directly the image caption (more details in Method section 4.1). (b) The distribution of covered body parts or cancer types in our constructed instruction data. (c) The distribution of question-and-answer sentence lengths in our instruction data. (d) The workflow of CLOVER. CLOVER employs the training framework of BLIP-2 to achieve a fast domain tuning with lightweight parameters. The entire training process of CLOVER includes two major stages: (i) alignment of vision and language and (ii) supervised fine-tuning with instructions. The alignment compels the model to acquire valuable representations between vision and language. Instruction fine-tuning is vital here for activating LLMs to excel in visual language question answering. Stage 1 requires inputs of image-text pairs, where we use the large-scale Quilt-1M dataset. Stage 2 demands our self-constructed domain-specific instruction data.}
    \label{Architecture}
\end{figure}

\section{Results}
\subsection{Model Overview}
A schematic illustration of the CLOVER is offered in Fig. \ref{Architecture}. To achieve fast domain tuning with low training resources, we adopt the BLIP-2 architecture \cite{blip2} as a visual language pre-training using a lightweight trainable query transformer (Q-former), a frozen visual encoder, and a frozen LLM. We leverage the paired pathological image and text captures from the Quilt-1M \cite{Quilt-1M} dataset to align the vision and language. For the instruction fine-tuning, we propose two approaches to generation instruction data (Fig. \ref{Architecture}(a)). We meticulously design prompts tailored for pathological question answering and use GPT-3.5 \cite{chatgpt} for generating effective instruction dataset at a low cost, referred as the ``generation-based instructions". The instructions generated by our PVLM-orient prompt show strong capabilities in the model tuning. Additionally, we construct instructions by matching template questions with the original text captions to enhance the model's vision understanding ability, referred as the ``template-based instructions". See more details in Methods section. Integrating these two types of instructions gives rise to the hybrid-form instructions, which remarkably enhance CLOVER's conversational abilities in pathology. Using two benchmark datasets and one independent clinical dataset, we comprehensively validate the effectiveness of CLOVER and demonstrate its potential for assisting clinical pathological tasks by ``talking to your pathology data" in resource-constrained settings. We discuss the related literature to CLOVER in Supplementary Related Works section ).

\subsection{Quantitative Comparison}
We systematically measure CLOVER's VQA performance using two public benchmark datasets, namely PathVQA \cite{pathvaq} and QUILT-VQA \cite{Quilt-LLaVA}, and one independent clinical data set (see Sec. Datasets). We compare CLOVER's performance with state-of-the-art (SOTA) PVLMs trained with notable costs to observe the cost-effectiveness of CLOVER (see Sec. Compared Methods and Instructions for more details about other PVLMs).
 
From Table \ref{sota}, CLOVER outperforms major competing methods on PathVQA dataset, showing a remarkable improvement of 26.13\% in accuracy at maximum for closed-ended questions. Our model performance is even approaching to LLaVA-Med (37 times more model parameters) with a minor accuracy difference of 1.83\%. Note that LLaVA and LLaVA-Med achieve their results through extensive parameter tuning of LLM and training on a vast instruction dataset generated by GPT-4. To illustrate, our method involves training with a frozen LLM, adjusting only a small portion of parameters (about 1/37 of LLaVA-Med). Meanwhile, CLOVER only involves the use an entry-level GPT-3.5 with a smaller scale of instruction dataset (2/3 of LLaVA-Med). As seen in Table \ref{sota}, our model also shows superior performance improvement in open-ended question scenarios. CLOVER's performance is twice that of BLIP-2 and approaches five times that of LLaVA in open-ended question scenarios. Compared with BLIP-2, our model improves in both closed-ended and open-ended question-answering. This key finding suggests that vision-language models could evidently benefit from the high-quality pathology-sensitive instruction data towards a low-cost model development.

We validate CLOVER's zero-shot generalization capability on the QUILT-VQA dataset. From Table \ref{QUILTVQA}, CLOVER based on GPT-3.5 (default setting) achieves the leading performance in precision and F1-score. Our results are even close to LLaVA-Med and Quilt-LLaVA in terms of recall. Due to the longer answer length of LLaVA-Med and Quilt-LLaVA, these models are advantageous in the recall evaluation metric. The average length (word counts) of true answers is 18 and that number for LLaVA-Med and Quilt-LLaVA are 36 and 55 respectively, both of which are higher than BLIP-2 and our model (4 and 22 respectively). In addition, our precision exceeds LLaVA-Med's by 10.25\% and Quilt-LLaVA's by 18.82 \%, while our F1-score also surpasses LLaVA-Med's by 6.53\% and Quilt-LLaVA's by 14.19\%. In terms of a performance-cost ratio (the ratio of performance to the log of the number of parameters), CLOVER based on GPT-3.5 achieves the leading values for all metrics. CLOVER based on LLaMA-3.1 \cite{llama3} and GPT-4o-mini also offer interesting advances. Overall, these findings indicate the strength of CLOVER to retain the high-level performance while minimizing costs.

\begin{table}[!t]
\centering
\resizebox{\columnwidth}{!}{
\begin{tabular}{ccccccc}
\toprule
Method & \small {\begin{tabular}[c]{@{}l@{}}Number of \\ instruction \end{tabular}} & \small{\begin{tabular}[c]{@{}l@{}}Trainable \\ parameter (pt) \end{tabular}} & Closed-end &   \begin{tabular}[c]{@{}l@{}}Closed-end \\ / log(pt) \end{tabular}    & Open-end & \begin{tabular}[c]{@{}l@{}} Open-end \\ / log(pt)\end{tabular} \\
\midrule
\small {VL Encoder–Decoder}  & N/A & 400M  & 84.63   & 32.52 &     -    & -  \\
Q2ATransformer   & N/A    & N/A  & 88.85    & - &      -    & - \\
M2I2   &    N/A           & 236M  & 88.00      & {\ul 37.09}  &      -     &  -\\
LLaVA    & 158K            & 7B & 63.20   & 16.41  & 7.74    & 2.01 \\
LLaVA-Med    & 60K       & 7B & \textbf{91.21}  & 23.68  & \textbf{37.95}  &{\ul 9.85}  \\
BLIP-2      &  N/A          & 187M & 83.90   & 36.93  & 18.69    & 8.23 \\
CLOVER (ours)       & 45K            &187M & {\ul 89.38}  & \textbf{39.34}  & {\ul 36.95}     &\textbf{16.26}\\
\bottomrule
\end{tabular}
}
\caption{Comparison with SOTA methods on PathVQA dataset. Bold and underline indicates the best and second best performances, respectively.}
\label{sota}
\end{table}

\begin{table}[!t]
\centering
\begin{tabular}{ccccccc}
\toprule
        
      Method                            & Recall  & \begin{tabular}[c]{@{}l@{}}Recall \\ / log(pt) \end{tabular} & Precision & \begin{tabular}[c]{@{}l@{}}Precision \\ / log(pt) \end{tabular} & F1-score  &  \begin{tabular}[c]{@{}l@{}}F1-score \\ / log(pt) \end{tabular} \\
\midrule
LLaVA                         & {\ul 57.36}  & 14.89 & 30.97 & 8.04   & 36.83 & 9.56 \\
LLaVA-Med                         & 57.03 & 14.81  & 30.49 & 7.92  & 37.03 &9.62 \\

Quilt-LLaVA & \textbf{59.95} & 15.57  &21.92 & 5.69 &29.37 & 7.63\\
BLIP-2                          & 10.03  & 4.41 & 30.51  & 13.43  & 13.95 & 6.14\\

CLOVER (LLaMA-3.1 70B) &45.07  &{\ul 19.84} &{\ul 38.04} & {\ul 16.74}&{\ul 38.68} &{\ul 17.03} \\

CLOVER (GPT-4o-mini) & 42.93 &18.90 &36.51 & 16.07&36.99 &16.28 \\
CLOVER (GPT-3.5)     &  54.33  & \textbf{23.91} & \textbf{40.74} & \textbf{17.93}   & \textbf{43.56} & \textbf{19.17}\\
\bottomrule
\end{tabular}
\caption{Comparison with prior SOTA methods on QUILT-VQA dataset. Bold and underline indicates the best and second best performances, respectively.}
\label{QUILTVQA}
\end{table}

\subsection{Qualitative Comparison}
To gain insight into model outputs, we present the representative cases from VQA experiments involving comparisons with LLaVA, LLaVA-Med, Quilt-LLaVA, BLIP-2, and our CLOVER on QUILT-VQA and LLaVA-Med-Pathology datasets. In Supplementary Table 1 for results from QUILT-VQA, we observe that the output of LLaVA is unrelated to the actual content, reflecting general image contents such as ``people", ``water", and ``lake". While LLaVA-Med identifies a tissue sample and provides definitions for pathology and immunohistochemistry, it does not reveal the specific type contexts of tissues. Quilt-LLaVA does not correctly recognize the image as a cross-section of bone. Instead, it mistakenly identifies it as a histopathological section and inaccurately suggests a possibility of cancer. BLIP-2 could only provide a simple answer without specific information. In contrast, CLOVER identifies that the image shows a compositional section of bone tissue and describes the composition of bone tissue and the structural shape, which serves as the basis for model reasoning. In addition, we perform evaluation on the LLaVA-Med-Pathology instruction. It is noted that results of LLaVA-Med tend to be overly optimistic due to the known overlap with its own training set \cite{llava-med}. A case is shown in Supplementary Table 2. LLaVA's answers still remain unrelated to pathology, while BLIP-2 could only answer generally without image details. Quilt-LLaVA provides incorrect responses regarding the number and positions of arrows and fails to deliver specific answers. Our model describes these inflammatory cells and provides a more detailed description of specific types of inflammatory cells. This is facilitated by our proposed PVLM-oriented prompts and generation-based instructions (Fig. \ref{Architecture}(a)) enriching the pathology knowledge distilled from GPT-3.5. Overall, CLOVER demonstrates its descriptive ability in reasonably responding to professional queries related to tissue characteristics and pathological outcomes. More qualitative comparisons are offered in Supplementary Case Study section, where the observations remain similar.

\begin{figure}[!t]
\centerline{\includegraphics[width=0.8\columnwidth]{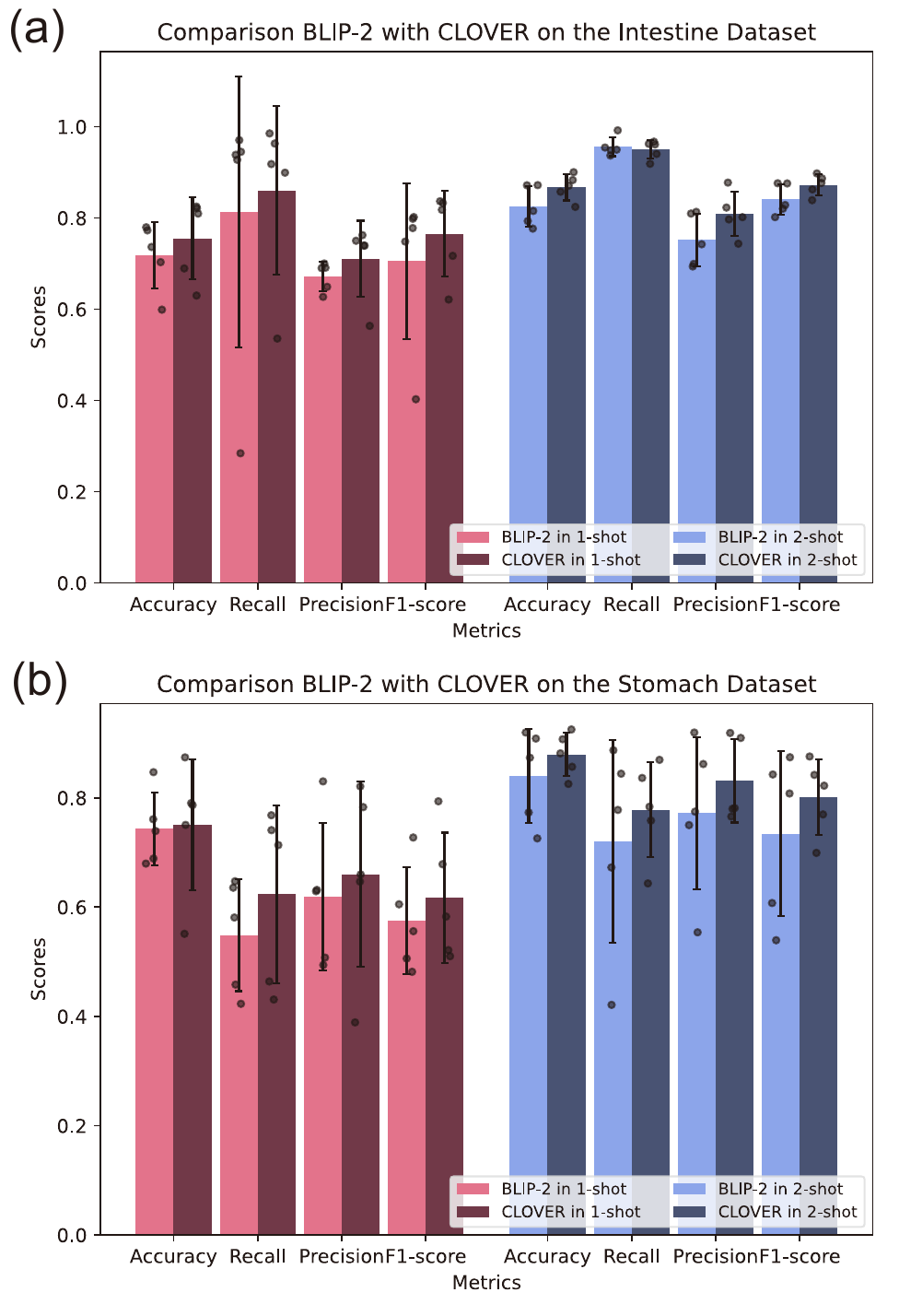}}
\caption{Comparison with prior SOTA methods on the intestine and stomach datasets. Each experiment was performed five times with different training samples. The bars present the mean values, and error bars present the standard deviations.}
\label{sto_int}
\end{figure}

\subsection{External Clinical Data Validation}
Few-shot learning enables models to make accurate predictions with a limited amount of labeled data. This setting allows a rapid adaptation to new clinical scenario without heavy labor and enables applications to rare diseases where clinical data is sparse and often difficult to access. We validate CLOVER's few-shot learning capability under a challenging task of cancer detection (cancer/non-cancer tissue classification). We implement a K-shot learning testing scheme, meaning that our model can only use K WSIs (K = 1, 2) from each class for a model fine-tuning. We evaluate the model performance on the external validation dataset on two cancer tissues, including intestinal and gastric cancer detection. We ensure the model has never seen the samples except the given shot samples (see Methods). Fig. \ref{sto_int}(a) details the performance of CLOVER in the intestinal cancer detection. We report comparative results from BLIP-2 model as it is not feasible for a RTX 3090 to fine-tune resource-demanding LLaVA-like methods. In addition, Fig. \ref{sto_int}(b) presents results in the gastric cancer detection. Overall, both Fig. \ref{sto_int}(a) and \ref{sto_int}(b) show that CLOVER demonstrates a performance advantage over BLIP-2. Even under the extreme condition of 1-shot learning, CLOVER achieves a commendable accuracy of 75.49\% in the intestinal classification task in Fig. \ref{sto_int}(a). In the gastric cancer detection, CLOVER's performance exhibits a rapid improvement with the increase of training samples. Starting from an accuracy of 75.04\% in the 1-shot scenario, the accuracy swiftly increased to 87.91\% with 2-shot learning in Fig. \ref{sto_int}(b). This marked improvement underscores CLOVER's potential on few-shot learning capabilities in cancer detection on the both stomach and intestine cancer tissues, especially under extreme sample conditions.

\begin{table}[!t]
\centering
\begin{tabular}{ccccc}
\toprule
LLM       & Instruction data &Estimated cost & Closed-end & Open-end \\
\midrule
FlanT5XL  & LM (from GPT-4)  & more than \$1,000            &  86.49     &  24.55    \\
FlanT5XL  & LM-IM (from GPT-4)  & more than \$1,000          & {\ul 86.56}    & 23.64    \\

FlanT5XL  & Quilt-instruct (from GPT-4)  &  \$8,804          & \textbf{87.67}     & {\ul 25.00}   \\

FlanT5XL  & Ours (from LLaMA-3.1 70B)  &  free          & 86.25     & 24.96   \\

FlanT5XL  & Ours (from GPT-4o-mini)  &  \$4         & 83.90     & 22.20  \\

FlanT5XL  & Ours  (from GPT-3.5)   & \$8          &  85.84     & \textbf{26.77}     \\
\midrule
Vicuna 7B & LM (from GPT-4)     & more than \$1,000           & 88.73          & 35.28         \\
Vicuna 7B & LM-IM (from GPT-4)   & more than \$1,000          & {\ul 89.88}     &  35.48    \\

Vicuna 7B &  Quilt-instruct (from GPT-4)  &  \$8,804          & 88.79     & 35.44  \\

Vicuna 7B  & Ours (from LLaMA-3.1 70B)  &  free          &89.09      &33.40    \\

Vicuna 7B  & Ours (from GPT-4o-mini)  &  \$4         & \textbf{90.39}     & {\ul35.99}  \\
Vicuna 7B & Ours (from GPT-3.5)  &  \$8      & 89.38     & \textbf{36.95}   \\
\bottomrule
\end{tabular}
\caption{The VQA performances of different models tuned with SOTA instruction datasets and our instructions. Bold and underline indicates the best and second best performances, respectively.}
\label{LM}
\end{table}
\subsection{Ablation Studies}
We conduct ablation studies on assessing the value of instruction data. We compare our instruction data with SOTA instruction datasets using the same model setting using BLIP-2 and use PathVQA as the final testing set. For CLOVER, we use GPT-3.5 (default setting), GPT-4o-mini, and an open-source LLaMA-3.1 to generate instructions, respectively. And we use the generated 15k generation-based and 30k template-based instructions as our instruction data. In Table \ref{LM}, the gains from using our instructions generated by GPT-3.5 based on noisy internet data are higher compared to using instructions generated by GPT-4 based on high-quality data. This finding can be attributed to the developed form of the instructions (generation-based and template-based instructions) and the quality of the instructions (generated through PVLM-oriented prompts). In particular, our costs based on GPT-3.5 are only \$8, whereas the estimated API costs of LLaVA-Med and Quilt-instruct \cite{Quilt-LLaVA} are hundreds of times higher than our low-cost approach. When using the FlanT5XL model, our performance on closed-ended question-answering closely approaches that of the LLaVA-Med-Pathology-IM (LM-IM) \cite{llava-med} and Quilt-instruct, evidently outperforming LLaVA-Med-Pathology (LM) \cite{llava-med} and LM-IM on open-ended question-answering by 2.22\% and 3.13\% respectively. With the Vicuna 7B model, our results surpass other instructions on both closed-ended and open-ended question-answering settings. Meanwhile, CLOVER with GPT-4o-mini offers certain advantages in close-end tasks, costing even lower prices. CLOVER with LLaMA-3.1 is competitive in close-end tasks. CLOVER with GPT-3.5 is still desired for open-end tasks. Overall, our results suggest that utilizing GPT-4 does not necessarily lead to a substantial performance improvement. CLOVER represents an alternative approach to achieve high performing PVLM particularly with low cost.

\begin{figure}[!t]
\centerline{\includegraphics[width=0.8\columnwidth]{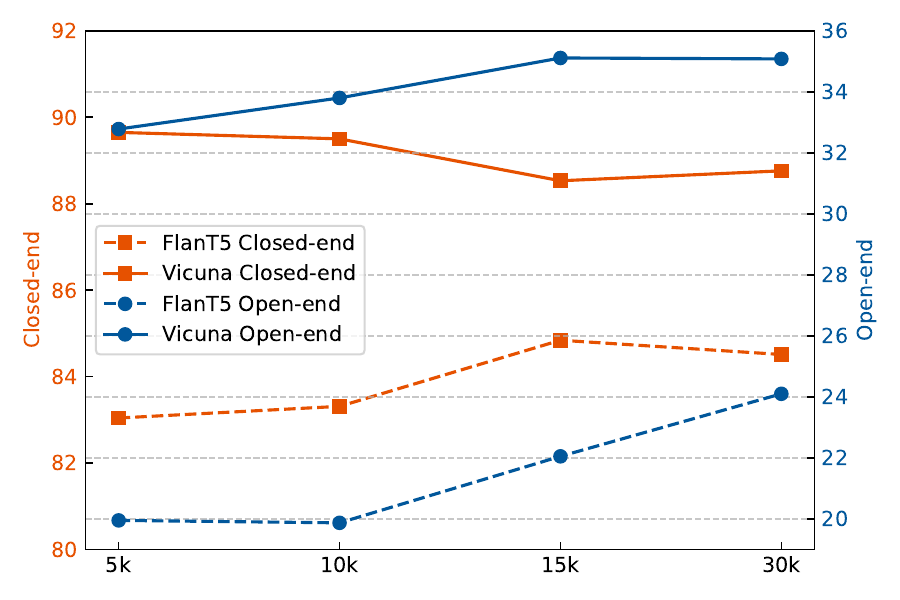}}
\caption{Model performance at different scale of generation-based instruction data on the PathVQA dataset. The trends of different methods are depicted with curves in different colors and dot shapes.}
\label{different-scale}
\vspace{-0.3cm}
\end{figure}\
Due to the expensive cost in the use of GPT API and high training expense, our goal is to generate less instruction data while retaining a high performance of PVLM. We are particularly focused on measuring the model performance in terms of a low demand of instruction data. We conduct experiments using 5K, 10K, 15K, and 30K generation-based instruction data for the FlanT5XL and Vicuna models. As seen in Fig. \ref{different-scale}, our method achieves high-quality results on small-scale instruction data. When comparing the cases of 15K and 30K instructions, the result difference is not statistically significant. For instance, the p-value for Vicuna on open-ended tasks is 0.1934 (two-sided t-test, t-statistic=-1.4201), and the p-value for closed-ended tasks is 0.4002 (two-sided t-test, t-statistic=-0.8885). This indicates that large-scale instruction datasets may not always be instrumental. More critically, we recognize that the form of the instructions (whether they possess template-based instructions) and the quality of the instructions (whether they are generated with high quality through PVLM-oriented prompts) are more differential factors compared to the quantity of instruction data.

We investigate the impact of different ratios between instructions on the performance. In this experiment, Vicuna is used as the LLM, and the total number of instructions is controlled to 20K. For PathVQA, in Supplementary Table 3, we find that closed-ended results monotonically improve as the ratio of generation-based instructions increases, while open-ended results improve as the ratio of template-based instructions increases. These results using QUILT-VQA are shown in Supplementary Table 4. The results demonstrate that the model's recall and F1-score improve progressively as the proportion of generation-based instructions increases. These these observations are consistent with above analysis. The generation-based and template-based instructions could be complementary to boost different model capacities in different datasets, and the ratio between them could be influencing. Note that these trends would not cover the results obtained by the CLOVER setting (15K generation-based and 30K template-based instructions), suggesting scales of the instruction sets are also crucial factors, to which we investigate in a following section.

To explore whether using simplified question-answering prompts can achieve results comparable to our designed PVLM-oriented prompts, we again utilize GPT-3.5 to generate instruction data but modify the prompt to be a single sentence. The defined prompt describes that ``Now you are a pathologist, and your task is to generate question-answer pairs based on the captions of the images." (examples are offered in Supplementary Table 5). The observable difference is that the answers of GPT-3.5 become shorter and lack additional knowledge. Results in Supplementary Table 6 show that for the relatively weaker-performing FlanT5XL, low-quality prompts have a serious impact, yielding a recall of only 12.64\% in open-question scenarios, even lower than BLIP-2's 18.69\%. This finding indicates that our PVLM-oriented prompts play a crucial role in enriching the instruction data with medical knowledge, thereby contributing to the performance improvement of the pathological vision-language model.

To explore the impact of epochs at each training stages, we conduct ablation experiments using Vicuna as LLM in Stage 1 (alignment), Stage 2 (instruction tuning), and the fine-tuning stage. We use the same model parameter initialization of BLIP-2. The experimental results are shown in Supplementary Figure 1 indicating that sufficient training on pathological images in Stage 1 (Supplementary Figure 1(a) and (b)) and Stage 2 (Supplementary Figure 1(c) and (d)) is necessary for BLIP-2's adaptation to the pathological domain. In the fine-tuning stage (Supplementary Figure 1(e) and (f)), increasing the number of epochs from 10 to 20 leads to a noticeable performance improvement in open-question tasks, as the model needs to adapt to the evaluation dataset according training dataset.

The quality of the generated instruction data could vary across different subsets. To validate the homogeneity and robustness of the instruction dataset, we randomly divide the 15K generation-based instruction data into three subsets of 5K each and use them for training separately. From results in Supplementary Table 7, we observe a stable performance across subsets (two-sided ANOVA, open-ended: p-value=0.2598, F-statistic=1.5113; close-ended: p-value=0.1179, F-statistic=2.5685), which indicates a potential generalization ability and adaptability to variations within the dataset. This homogeneous ability is critical for measuring the robustness of instruction data since variations and noises can be introduced during data collection.

\section{Discussion}
Computational pathology is widely known for its high demands on data and computational resource, especially for building a useful PVLM application. Prior efforts were primarily focused on the use of human-examined clinical data \cite{zhang2024data, clinical1}. A key differentiation of our study is to build efficient PVLMs based on the public Internet data and align with the power of instruction data without extensive human examinations. To address the challenge of image-to-text alignment, we show that using the generation-based instructions and prompting with GPT-3.5 can generate high-quality instruction data. In addition, we offer template-based instructions to improve the scale and diversity of the instructions, leading to the improved generalizability of models in pathology. It is noteworthy that combining GPT-3.5 and proper prompt could result in a high performance that even outperforms the results from GPT-4. Since we reiterate the performance gain without advanced GPTs, our study can serve as a methodological baseline for guiding follow-up studies to measure cost-effectiveness.

Data generation has become increasingly crucial to diversify training data and enhance model robustness \cite{chang2023mining, graikos2024learned, ding2023large, pathgen}. In our study, instruction data generation is a core strategy for CLOVER to achieve its strong efficiency. Especially in the era of large model, instruction data generation has proven to be useful in model fine-tuning as it can unlock the specialized use of LLMs \cite{instruction-fintuned1, instruction-fintuned2, instruction-fintuned3}. From our results, we have gained key insights into an efficient use of instruction in model tuning and inference. First, instruction fine-tuning markedly improves performance compared to models without fine-tuning. Even under low-resource conditions with frozen LLM parameters, instruction fine-tuning can still enhance the performance. Second, small-scale, well-defined instructions are powerful on guiding PVLM inference, suggesting that large-scale instruction datasets are not always necessary. This finding is strongly aligned with the prior research that less but higher quality instruction data can yield superior results \cite{Less-is-more}. Our well-crafted prompts are more advantageous for a LLM to generate high-quality instructions when comparing the distinction of PVLM-oriented prompts versus non-PVLM-oriented prompts on model performance. Finally, our study confirms that proper use of instructions is impactful in complex pathological visual question answering tasks. This key insight supports the observation that instruction tuning is more beneficial for complex and unseen tasks compared to simpler ones \cite{wei2021finetuned}.

It has come to our attention that instructions generated using PVLM-oriented prompts naturally incorporate the characteristic of chain-of-thought (CoT) \cite{CoT}. From our results, these instructions typically combine explicit answers with explanations and contextual expansions on the medical knowledge. In addition, previous investigations have suggested that combining non-CoT instructions with CoT instruction fine-tuning achieves positive results, outperforming the solely CoT instruction fine-tuning \cite{instruction-fintuned1}. Likewise, our finding affirms that a hybrid-form of instructions, encompassing both generation-based and template-based instructions, can greatly contribute to the development of cost-efficient PVLMs.

Our primary focus has placed on building a low-cost PVLM, we thus have not address the automated noise removal and fine-scale image-to-text alignment that can potentially improve the model robustness performance. CLOVER studies patch-level pathology image processing due to its computational efficiency and does not involve WSI-based application \cite{Wsi-vqa}, thus the extensive analysis on spatially-aware image contents can enhance clinical diagnosis and report generation \cite{ding2022spatially}. While our study simultaneously aims to sweep the barrier in data, computational source, and financial costs for building PVLM, other paralleled cost-effective endeavours \cite{lora, jin2024efficient} can be considered to enhance performance efficiency. As our instruction approach can potentially extend to other demanding human-AI interactions, the continued exploration of external multi-modal clinical data could help validate the low-cost utility of CLOVER. Finally, our study demonstrates the power of instruction data for building a vision-language model, enabling the rapid language-and-image information interaction and conversational decision making in the space of digital pathology.

\section{Methods}

In this section, we offer methodological details of the development of CLOVER. We next introduce the involved datasets and the evaluation schemes of experiment.

\subsection{Efficient Instructions Construction at Low-cost}
Generating large-scale instructions via GPT-4 incurs substantial financial costs. In this study, we specialize in developing effective domain-specific instruction data generation at a low cost. The proposed framework includes (i) generation-based instructions with a specialized prompt for employing GPT-3.5, and (ii) template-based instructions without any additional financial cost. The construction of generation-based instructions involves a generation of question-answer (QA) from the captions using GPT and a PVLM-oriented prompt, while that of template-based instructions involves matching the captions as answer to a set of pre-designed template questions. Notably, the template-based QA dataset permits a comprehensive understanding of image contextual information, while the generation-based QAs emphasize the pathological knowledge distilled from GPT-3.5.

\textbf{Generation-based Instruction.} We meticulously design a prompt tailored for pathological question answering. We use GPT-3.5 \cite{chatgpt} for instruction dataset generation enabling a low cost operation with PVLM-oriented prompt, as seen in Supplementary Table 8. Given the high variance of prompt design, we have designed four desired principles of the prompt construction. First, we use GPT-3.5 to simulate a scenario where users (patient or doctor) and AI assistants (CLOVER model) conduct question and answering (QA). Since GPT-3.5 does not have access to images, QAs generated by GPT-3.5 are based on the textual description of the image. To reduce the over-reliance on textual descriptions, we emphasize avoiding minor information that can not be obtained from the image (for instance, reference dates or magnification ratios). Second, we focus on adding visual detailed information such as tissue structure, cell morphology, potential lesions, and locations. Third, the noise in the original textual description is avoided such as vocabularies related to context or narrator. Finally, we seek to generate answers of GPT-3.5 to exhibit the cautiousness, aligning with the medical field's expectation for producing prudent answers. To enhance the quality of generated data, we additionally introduce few-shot examples in the prompts to inject more relevant information for in-context learning \cite{in-context}.

\textbf{Template-based Instruction.} The above process of generation-based instruction with GPT-3.5 is often based on partial captions derived from images. Generation-based instructions only capture a portion of the original information, preventing LLMs from fully capturing the visual and descriptive content. To address this challenge, we construct useful template-based instructions, providing the LLM with more structured and comprehensive language guidance, overcoming the shortcoming where visual information are often overlooked or incomplete in pathology. We use the same 17 descriptive statements as LLaVA \cite{llava}, which instructs the model to intricately describe the content of images using various expressions. These statements are the questions and the corresponding answers from the original captions of the images. In detail, we merge multiple captions for the same image and filter out those with a word count less than 25. Subsequently, we randomly choose 30K image-text pairs for generating the template-based instruction. For each image-text pair, we randomly select only one description from 17 statements to form our template-based instructions. Note that this generation process requires no additional fee as the use of GPT is not involved.

\subsection{Training Details of CLOVER}
Model training involves two main stages: (i) alignment of vision and language and (ii) supervised fine-tuning with instructions (Fig. \ref{Architecture}(d)). In the first training stage, to align pathological images and text, we train BLIP-2 on the original image-text pairs directly obtained from Quilt-1M dataset \cite{Quilt-1M}. BLIP-2 takes inputs in the form of a pair of image and text (caption or question). The visual encoder is utilized for extracting features from image and generating visual tokens $\textbf{V} = \{v_1, v_2, ..., v_{n_v}\}$, where $n_v$ is the number of visual tokens. Next, the lightweight Q-former handles text and visual learnable queries, incorporating a self-attention \cite{attention} mechanism to share the transformer for both visual and textual components. A cross-attention mechanism is used to facilitate the interaction between visual tokens and visual queries. We simultaneously optimize the Q-former using image-text contrastive loss, image-grounded text generation loss, and image-text matching loss \cite{blip}. The image-text contrastive loss aims to maximize the similarity between the same image-text pair by comparing visual queries and text representations. The image-grounded text generation loss trains a text transformer to generate corresponding text given an image. Meanwhile, the image-text matching loss is a binary classification loss used to determine whether an image and text belong to the same pair.

In the second stage training, we introduce the customized instruction dataset for activating LLM and completing visual language question answering. In order to stimulate the domain speciality of LLM, we add a task-driven prompt before the input question as ``Now that you are a pathologist, please answer the following questions based on the images". We utilize a standard tokenization to obtain a sequence of text tokens, including prompt tokens $\textbf{P} = \{p_1, p_2, ..., p_{n_p}\}$, question tokens $\textbf{Q}= \{q_1, q_2, ..., q_{n_q}\}$, and answer tokens $\textbf{A} = \{a_1, a_2, ..., a_{n_a}\}$, where $n_q$, $n_p$ and $n_a$ represent the token lengths of prompt, questions and answer. We feed the Q-former both the visual tokens $\textbf{V} = \{v_1, v_2, ..., v_{n_v}\}$ and the text tokens, and then adjust a linear layer to map the dimension of image token into the the dimension of text. Our optimization objective is to fine-tune the learnable parameters $\theta$ of Q-former by maximizing the following likelihood:
\begin{equation}
\begin{aligned}
    p(A|P,Q,V) = \prod_{i=1}^{n_a}  p_{\theta}(a_i|P,Q,V, a_1, a_2, ..., a_{i-1}),
\end{aligned}
\end{equation}
where $p_{\theta}(a_i|P,Q,V, a_1, a_2, ..., a_{i-1})$ is the probability to generate $a_i$ given $P$, $Q$, $V$,  $a_1$, $a_2$, ..., and $a_{i-1}$ under the model parameters $\theta$.

\subsection{Implementation Details of CLOVER}
We choose EVA-ViT-G/14 \cite{evavit} as the frozen visual encoder, utilizing the output from its second-to-last layer as the visual feature representation. This choice has been validated as a superior solution in BLIP-2 \cite{blip2}. Regarding the LLM, we select the decoder-only Vicuna 7B \cite{vicuna} and encoder-decoder based FlanT5XL \cite{flant5}, which could best utilize the pre-trained BLIP-2 parameters. To enhance the training efficiency, we convert the parameters of visual encoder and LLM to FP16. In the first stage of model training, we conduct training for 20 epochs with a batch size of 36. In the second stage, training continues for 30 epochs, with a batch size of 2 for Vicuna and 8 for FlanT5XL. The remaining hyperparameter settings remain consistent with BLIP-2. We complete the two-stage training using Vicuna in 4 days with 4 RTX-3090 GPUs.

\subsection{Datasets}
\subsubsection{CLOVER Training Dataset}
\textbf{Quilt-1M} \cite{Quilt-1M} is a multi-modal pathology dataset involving both pathological vision and language information. Quilt includes 768,826 pathological images and their corresponding textual annotations. The images are screen shots of educational histopathology YouTube videos from expert clinicians, and the text annotations are extracted based on the corresponding audio to the image. The text annotations are further filtered and refined by a LLM with prompts. Based on Quilt, additional established Internet-based (for instance, Twitter and research papers) image-text paired data are integrated to form a total dataset of one million image-text pairs, called Quilt-1M. In our study, we use Quilt-1M in two tasks including the first-stage vision-language alignment and instruction generation for supervised fine-tuning.

\textbf{CLOVER instruction} is generated using Quilt-1M dataset with our proposed method as introduced above. It consists of the generated 15k (a default size in major experiments) generation-based using GPT-3.5 and 30k template-based instructions based on manual construction. We develop this original instruction dataset for activating LLMs to complete visual-language question answering in pathology domain. These instructions are used for instruction tuning (Stage 2). The generated CLOVER instruction covers a broad range of pathology VQA about different body parts and cancer types (as shown in Fig. \ref{Architecture}(b)), containing large-scale question and answer pairs with diverse complexity (as shown in Fig. \ref{Architecture}(c)). An example of generating question and answer instructions based on the prompt is shown in Supplementary Table 9.

\subsubsection{Evaluation Datasets}
\textbf{PathVQA} \cite{pathvaq} is a pathological visual question-answering dataset, comprising 4,998 pathological images and 32,799 question-answer pairs. The questions in this dataset are categorized into two types: open-ended questions and closed-ended questions. Open-ended questions cover a wide range of topics, including why, what, how, etc., while closed-ended answers are limited to responses like ``yes" or ``no". The training and testing splits are specified by the dataset. For experiments using PathVQA, the models are fined tuned by training dataset. Testing dataset is used for evaluation. This setting is consistent with previous studies \cite{llava-med} and provides the basis to compare our CLOVER with SOTA methods. Note that for all ablation studies, we use PathVQA by default. 

\textbf{QUILT-VQA} \cite{Quilt-LLaVA} is another pathological visual question-answering dataset. QUILT-VQA is uniquely sourced from educational video contents covering diverse topics. Researchers extract valuable texts from these videos. Then, GPT-4 is utilized to extract question-answering pairs from these texts with human intervention ensuring the pairs alignment on the medical themes. QUILT-VQA comprises 985 visual question-answer pairs (released before March, 2024), with an average word count of 17 in the answers. Again, to be consistent with previous work \cite{Quilt-LLaVA}, we use a zero-shot validation scheme in QUILT-VQA. In our study, we regard all questions as open-ended questions to fully utilize the dataset.

\textbf{Clinical dataset} is used to test CLOVER under a real-world clinical setting. We design a cancer detection task, where CLOVER performs in a visual question-answering (VQA) manner for the given image patch. We collect 38 cancer tissues WSI in Pathology Department of Xinhua Hospital affiliated to Shanghai Jiao Tong University School of Medicine, during April 2024 to May 2024, including 13 WSIs from stomach and 25 WSIs from intestines. To identify specific pathological regions within WSIs, we collaborate with two experienced pathologists to perform the fine-grained annotation with cross-checking. These annotations, provided in standard XML files, delineate the negative and positive regions within the WSIs. For testing CLOVER working on patch-level VQA, we further extract non-overlapping image patches (of size 512 $\times$ 512 pixels) containing the annotated sub-regions. This process yields 7,112 image patches (1,136 tumor and 2,079 non-tumor patches from stomach, 1,846 tumor and 2,051 non-tumor patches from intestines). For model training and evaluation, we formulate the cancer detection task into a visual question answering (VQA) format. The uniform question is posed as: ``Is this pathological image showing a negative or positive result?" The answers are set to ``this is a negative pathological image" or ``this is a positive pathological image" based on the ground-truth labels. For this experiment, we implement a few-shot learning setting to test whether our model can be fast transferred to the real-world data with a few annotated samples. We divide the data into training (15 WSIs) and testing sets (23 WSIs) on the WSI level, and the number of training WSIs depends on the setting of our five independent experiments. For both each 1-shot (patches of one random WSI from each class for training) and 2-shot (two WSIs from each class) experiment, we reconstruct the training samples using different combinations of samples from training samples. We ensure that there is no WSI data from overlapped patients. The test set remains constant throughout all experiments to facilitate the evaluation across different experiments. We collect the dataset from the private hospital and ensured that the testing data is never released in the Internet and not contained in the training set. Please note that we do not require the prediction class is new during the few-shot learning experiment.

\subsection{Compared Methods and Instructions}
We conduct a comprehensive comparison with the standard BLIP-2 and other strong baseline approaches, including VL Encoder-Decoder \cite{VLEncoder–Decoder}, Q2ATransformer \cite{q2atransformer}, M2I2 \cite{M2M}, LLaVA \cite{llava}, LLaVA-Med \cite{llava-med}, and Quilt-LLaVA \cite{Quilt-LLaVA}. Since VL Encoder-Decoder, Q2ATransformer and M2I2 do not have zero-shot generalization ability, these models are not considered when evaluating the performance on QUILT-VQA.

LLaVA-Med-Pathology, LLaVA-Med-Pathology-IM \cite{llava-med} and Quilt-instruct \cite{Quilt-LLaVA} are the public instruction datasets that we use to compare with our proposed instruction dataset. LLaVA-Med-Pathology is a high-quality conversational instruction dataset focused on the pathological domain and is a subset of LLaVA-Med \cite{llava-med}. LLaVA-Med-Pathology-IM is another version of LLaVA-Med-Pathology, which adds inline content to the original image descriptions as supplementary textual information. These datasets are created by transforming parts of the PMC-15M dataset \cite{pmc} into instruction datasets using GPT-4 based on specific prompts. Quilt-instruct also focuses on VQA in pathological domain. It extracts pathology images and captions from educational histopathology YouTube videos, with spatial localization based on narrators' cursor movements. GPT-4 is further introduced to refine the captions, offering contextual reasoning. Note that all these datasets are without manual verifications. Therefore, though we implement a qualitative case analysis with LLaVA-Med-Pathology, it does not support a quantitative analysis.

\subsection{Evaluation Metrics}
For PathVQA, we use the recall of the true answer that appears in the predicted answer in open-ended questions, and we report the accuracy in closed-ended questions. For QUILT-VQA, we find that under zero-shot learning, the answer lengths of different models vary greatly, and thus we report the recall, precision and F1-score. F1-score as an integration of recall and precision is used as a more comprehensive metric among the three. Precision is the proportion of correctly predicted words in the sentences generated by the model. Recall is the proportion of correctly predicted words in a standard answer. F1-score is the harmonic mean of precision and recall. The definition is as follows:

\begin{equation}
\rm   Precision = \frac{TP}{TP+FP},
\end{equation}
\begin{equation}
    \rm Recall = \frac{TP}{TP+FN},
\end{equation}
\begin{equation}
    \rm F1 = \frac{2*Precision*Recall}{Precision+Recall},
\end{equation}
where TP represents the number of words in common between the standard answer and the predicted sentence, FP indicates the number of words or characters in the predicted sentence but not in the standard answer, and FN indicates the number of words or characters that are in the standard answer but not in the predicted sentence. All above metrics are reported in percentage. To evaluate the cost-effectiveness of different models, we further compute the ratio of performance to the log of the number of parameters as a performance-cost ratio metric.

\subsection{Statistical Analysis}
To assess the statistical significance of the performance metric from single test dataset, we subsample the testing dataset into 5 equal folds and calculate the metrics within each sub-fold. The statistical comparison can then be conducted using the variations among these samples. To compare the performances under 15K and 30K instructions, we used two-sided t-test. To compare the performances among different instruction data subsets, two-sided ANOVA is applied. All analyses are implemented using SciPy toolbox in Python.

\backmatter

\section*{Data Availability}
The QUILT-1M, QUILT-VQA and Quilt-instruct \cite{Quilt-1M} can be accessed in \url{https://quilt1m.github.io/}. LLaVA-Med-Pathology \cite{llava-med} can be accessed in \url{https://github.com/microsoft/LLaVA-Med}. PathVQA \cite{pathvaq} can be downloaded from \url{https://huggingface.co/datasets/flaviagiammarino/path-vqa}. The clinical dataset from Xinhua Hospital is available upon request from the corresponding author (zhangshaoting@pjlab.org.cn) due to the privacy protection restriction of hospital. The request will be reviewed to ensure confidentiality. A data-sharing agreement must be signed prior to data release.

\section*{Code Availability}
The code, instruction datasets and models have been publicly available at \url{https://github.com/JLINEkai/CLOVER} and \url{https://doi.org/10.5281/zenodo.15081542}.

\section*{Acknowledgements}
This study is supported in part by Shanghai Artificial Intelligence Laboratory (ML and SZ), the Centre for Perceptual and Interactive Intelligence (CPll) Ltd under the Innovation and Technology Commission (lTC)'s InnoHK (SZ).

\section*{Author Contributions Statement}
KC, ML, MZ and SZ are major contributors to drafting and revising the manuscript for content and analyzing the data. FY, LM, XS, LW, XW, LZ, and ZW played major roles in the acquisition of data. KC, ML, MZ, FY, XS, LW, LM and XW substantially revised the manuscript. KC, ML, MZ and SZ conceptualize and design the study. ML, LZ and ZW interpret the data. All authors read and approved the final manuscript.

\section*{Competing Interests Statement}
The authors declare no competing interest.


\renewcommand{\thefigure}{\arabic{figure}}
\renewcommand{\thetable}{\arabic{table}}
\renewcommand{\tablename}{Supplementary Table} 
\renewcommand{\figurename}{Supplementary Figure} 
\setcounter{figure}{0}
\setcounter{table}{0}
\section*{Supplementary Related Works}\label{secA1}

The family of LLMs \cite{zhang2024data, flant5, vicuna, gpt4} and visual language pre-training \cite{clip, blip} are two indispensable building blocks for vision-language foundation models. Visual language pre-training aligns the encoding space in vision and language. The visual features from the visual encoder can then be perceived by the LLMs. Examples of multi-modal foundation models include Flamingo \cite{Flamingo}, BLIP-2 \cite{blip2}, FROMAGe \cite{FROMAGe}, mPLUG-Owl \cite{mplug-owl}, and LLaVA \cite{llava}. In particular, Flamingo, BLIP-2, and FROMAGe all freeze LLMs and allow vision to adapt to language information through different modules, while mPLUG-Owl and LLaVA all train LLM to fuse visual and language information. Their success strongly benefits from leveraging large-scale natural image-text datasets. For instance, LLaVA leverages a GPT-based data generation to construct the instruction dataset. Despite the promise shown in general domains, these approaches face inherent challenges in the healthcare system. This is due to the medical complexity that significantly differ from generic domains, requiring models to capture specialized disease concepts \cite{textguided, pathology-geo}. Therefore, extending these general models to the medical domain requires fine-tuning and the careful use of specific medical image-text pairs and question-answer data \cite{zhang2024data}.


Building medical vision-language foundation models requires systematic efforts in both data curation and model inference in the healthcare system \cite{zhang2024data}. The large-scale resources from the literature (PMC-15M \cite{pmc}), social media (PLIP \cite{Twitter} and Quilt-1M \cite{Quilt-1M}), accessible textbooks \cite{Med-flamingo} and private hospital data \cite{lu2024visual, lu2024multimodal} provide critical data support for domain-specific multi-modal models. Examples include PMC-LLaMA \cite{Pmc-llama}, LLaVA-Med \cite{llava-med}, and Med-Flamingo \cite{Med-flamingo}. LLaVA-Med is a representative work that aligns vision and language using 600K image-text pairs on the basis of LLaVA and fine-tunes the model using 60K dialogue-based instructions. Yet these instruction sets are generated by GPT-4 at a high cost. Similarly, in the pathological field, Quilt-LLaVA \cite{Quilt-LLaVA} uses Quilt-1M and GPT-4 to generate instruction data at a cost of up to \$8,858. Towards a better utility of foundation models, instruction preparation and generation in the various forms of structure and scale require substantial research efforts. Meanwhile, during the fine-tuning stage, LLaVA-like methods \cite{llava} update the parameters of the LLM, resulting in a high training cost that can not be accommodated on consumer-grade GPUs. This pressing demand for high-performance training equipment becomes a daunting hurdle for researchers to extend general-purpose models into domain-specific vision-language models. We adopt the BLIP-2 \cite{blip2} architecture as the foundation for vision-language pretraining. Discriminating from LLaVA-like models, BLIP-2 utilized a lightweight architecture and a pre-training on large-scale natural image-language data, which offers a firmed basis for CLOVER's vision-language and few-shot learning capacities, with a low computational requirement.

\section*{Supplementary Tables and Figures}\label{secB1}
The experimental results of CLOVER with different training epoch at different stages are shown in Supplementary Figure \ref{differentepochs}. Cases analyzed in main text can be found in the Supplementary Table \ref{QAQUILT-VQA} and \ref{QALLaVA-Med-17K}. The impacts of different ratios between instructions on the PathVQA and QUILT-VQA are shown in Supplementary Table \ref{description} and \ref{RatioQUILTVQA}. An example of the questions and answers respectively generated with PVLM-oriented prompt and non-PVLM-oriented prompt is shown in Supplementary Table \ref{QAcompared}, and the results are presented in Supplementary Table \ref{well-designed}. The results of the homogeneity and robustness of the instruction dataset are shown in Supplementary Table \ref{groups}. PVLM-oriented prompt is shown in Supplementary Table \ref{prompt}. An example of generating question and answer instructions based on the prompt is shown in Supplementary Table \ref{exampleQA}.

\begin{table}[!t]
  \centering
  \begin{tabular}{@{}p{0.1\linewidth} p{0.4\linewidth}p{0.4\linewidth}@{}}
    \toprule
    \textbf{Question:} &What can be observed in the image? & \multirow{3}{*}{\includegraphics[width=0.9\linewidth]{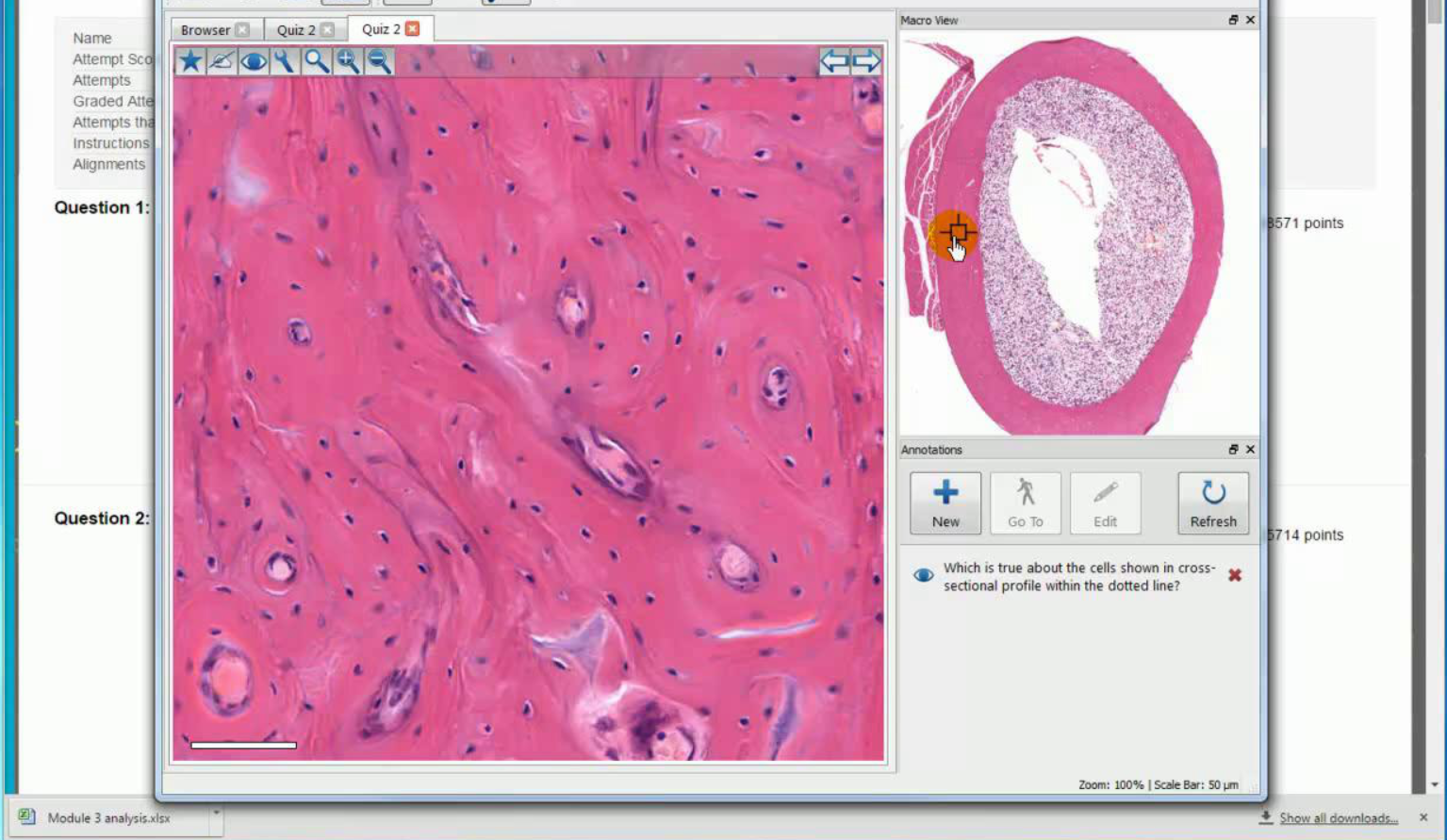}} \\
        \textbf{Answer:} &{\ul In the image, a cross-section of a bone can be seen upon magnification.} Osteomas in the compact bone are visible. The tissue in question shows a very eosinophilic cytoplasm and nuclei at the edge, indicating a cross-sectional profile of skeletal muscle cells.\\
    \midrule
    \textbf{LLaVA:}& \multicolumn{2}{p{0.8\linewidth}}{   In the image, there is a person standing in front of a large body of water, which appears to be a lake. The person is wearing a backpack, and there is a boat visible in the water. The scene suggests that the person might be taking a break from boating or preparing to go out on the water.} \\
    \textbf{LLaVA-Med:} & \multicolumn{2}{p{0.8\linewidth}}{ In the image, you can observe the histopathology and immunohistochemistry of a tissue sample. Histopathology refers to the microscopic examination of tissue to study the manifestations of disease, while immunohistochemistry is a technique used to visualize the presence and distribution of specific proteins or antigens in tissue samples using antibodies.}\\

    \textbf{Quilt-LLaVA:} & \multicolumn{2}{p{0.8\linewidth}}{The image appears to be a histopathological slide, possibly stained with Hematoxylin and Eosin (H\&E). The staining technique is commonly used in medical diagnosis to highlight different structures in tissue samples. The image shows a variety of cellular structures, some of which may be indicative of a pathological condition. The cells appear to be densely packed, suggesting a high degree of cellular proliferation. This could be a sign of a neoplastic process, but further examination would be required to confirm this.The nuclei of the cells are also visible, and they appear to be irregular in shape and size. This is a common feature of malignant cells, as the process of cell division becomes unregulated in cancer. The presence of these abnormal nuclei could suggest a diagnosis of cancer, but again, further examination would be required to confirm this.}\\
   \textbf{BLIP-2:} & \multicolumn{2}{p{0.8\linewidth}}{ A cell nucleus in a tissue.}\\
   \textbf{CLOVER:} &  \multicolumn{2}{p{0.8\linewidth}}{ {\ul Based on the provided image, it appears that the observer is looking at a section of bone tissue}, the bone tissue consists of a honeycomb-like structure made up of osteocytes, which are specialized bone cells embedded within the bone matrix. }\\
\bottomrule
  \end{tabular}
  \caption{Qualitative comparisons of visual question answering on QUILT-VQA. (Image source: QUILT-VQA \cite{Quilt-LLaVA})}
    \label{QAQUILT-VQA}
\end{table}

\begin{table}[!t]
  \centering
  \begin{tabular}{@{}p{0.1\linewidth} p{0.4\linewidth}p{0.4\linewidth}@{}}
    \toprule
    \textbf{Question:} &What are the arrows indicating? & \multirow{3}{*}{\includegraphics[width=0.55\linewidth]{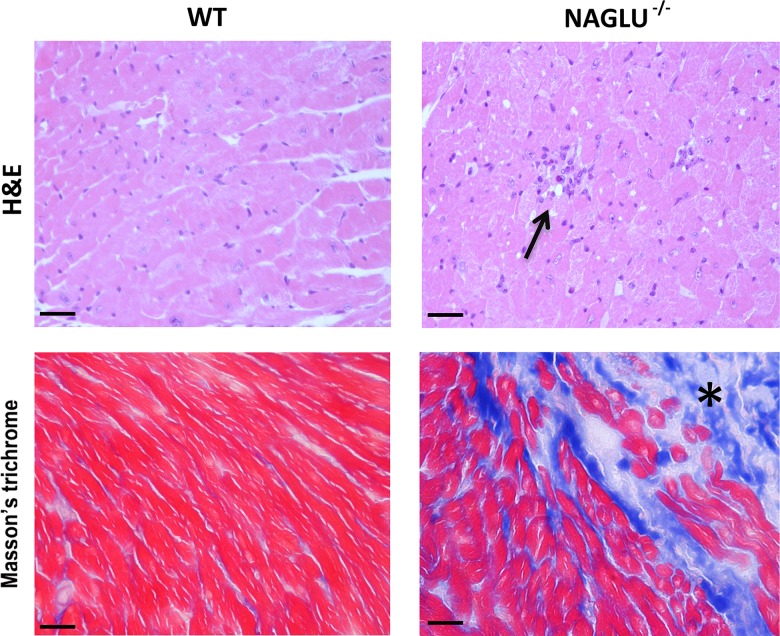}} \\
        \textbf{Answer:} & {\ul The arrows in the image are pointing to inflammatory infiltrates within the myocardial fibers.} Inflammatory infiltrates are collections of immune cells that have migrated to the site of inflammation or injury in the tissue.   \\
    \midrule
    \textbf{LLaVA:} & \multicolumn{2}{p{0.8\linewidth}}{  The arrows indicate the direction of the flow of energy in the system. In this case, the arrows point from the heat source to the heat sink, indicating that heat is being transferred from the heat source to the heat sink. This is a common way to represent the flow of heat in a system and it helps visualize how the system is operating and how it can be optimized for better performance.} \\
    \textbf{LLaVA-Med:} &\multicolumn{2}{p{0.8\linewidth}}{ The arrows in the image are pointing to areas of inflammatory infiltrate. Inflammatory infiltrate refers to the accumulation of immune cells, such as white blood cells, in the tissue as a response to injury, infection, or other factors that cause inflammation.}\\

    \textbf{Quilt-LLaVA:} &\multicolumn{2}{p{0.8\linewidth}}{ In the image, there are two arrows pointing towards the same area, which is located towards the upper left quadrant. The arrows are likely indicating a significant feature or abnormality in the tissue sample. However, without additional context or specific details about the histopathological features present in the image, it's challenging to provide a more precise interpretation.} \\
   \textbf{BLIP-2:}  & \multicolumn{2}{p{0.8\linewidth}}{The direction in which a tissue is moving.}\\
    \textbf{CLOVER:} & \multicolumn{2}{p{0.8\linewidth}}{{\ul The arrows in the image indicate the presence of inflammatory cells,} such as neutrophils and lymphocytes, within the myocardium these inflammatory cells are a hallmark of myocarditis, which is an inflammation of the heart muscle.}\\
\bottomrule
  \end{tabular}
\caption{Qualitative comparisons of visual question answering on LLaVA-Med-Pathology. (Image source:  \url{https://www.ncbi.nlm.nih.gov/pubmed/26147524})}
    \label{QALLaVA-Med-17K}
\end{table}

\begin{figure}[!t]
    \centering
    \centerline{\includegraphics[width=\columnwidth]{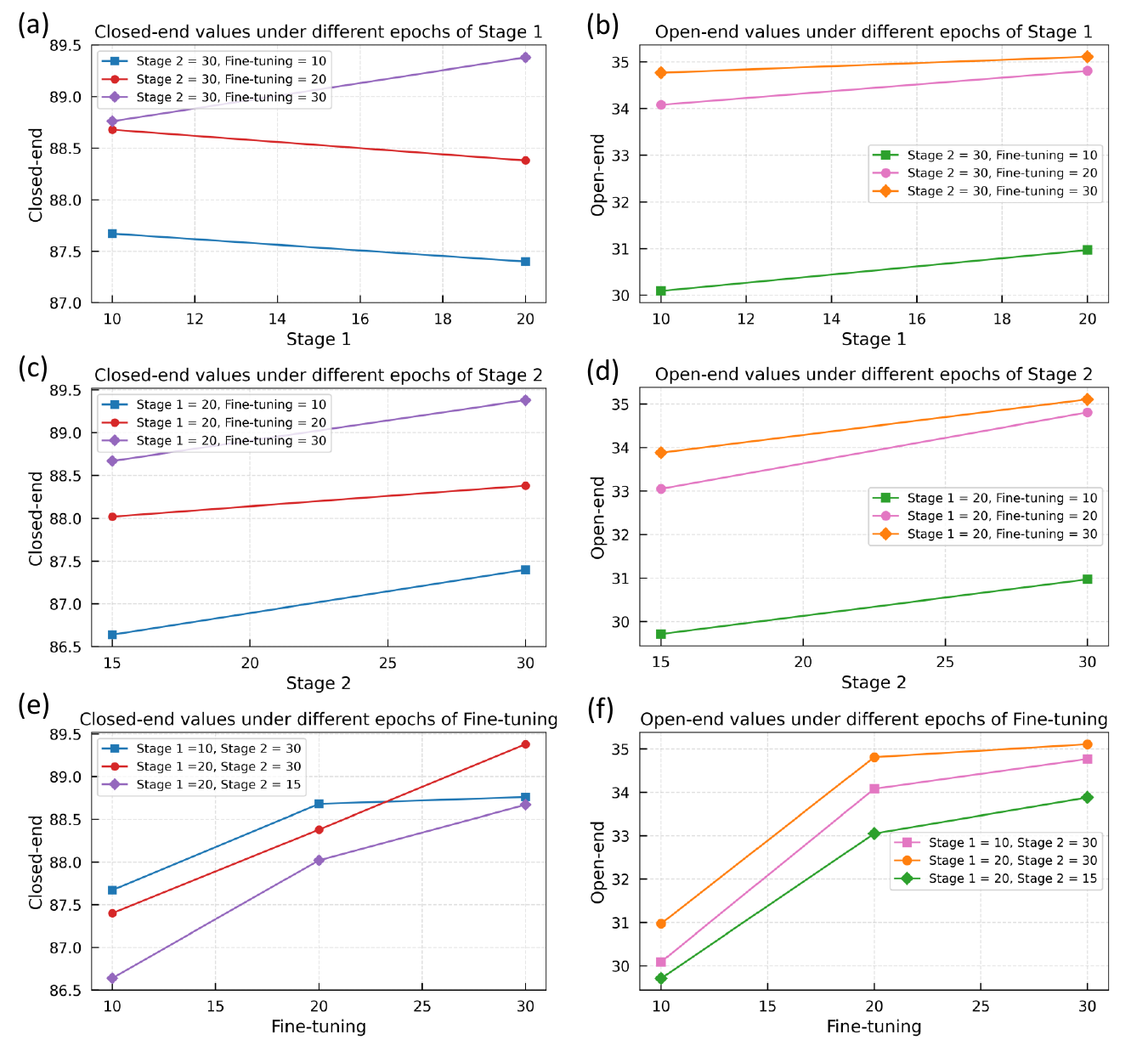}}
    
    \caption{VQA performance of CLOVER with different training epoch at different stages on the PathVQA dataset. (a)-(b). The influence of different Stage 1 (alignment) training epochs on closed-ended and open-ended VQA performances. (c)-(d). The influence of different Stage 2 (instruction tuning) training epochs on closed-ended and open-ended VQA performances. (e)-(f). The influence of different fine-tuning training epochs on closed-ended and open-ended VQA performances.}
    \label{differentepochs}
\end{figure}

\begin{table}[!t]
\centering
\begin{tabular}{cccc}
\toprule
LLM       & Instruction   & Closed-end & Open-end \\
\midrule
FlanT5XL  & 15K G + 30K T (CLOVER setting)          &  \textbf{85.84}     & \textbf{26.77}     \\
FlanT5XL  &  15K G           & 84.51     & 24.10     \\

FlanT5XL & 30K T               & 85.20    & 26.35  \\
\midrule
Vicuna 7B & 15K G + 30K T       & \textbf{89.38}     & \textbf{36.95}   \\
Vicuna 7B & 15K G               & 88.76     & 35.08   \\

Vicuna 7B & 30K T               & 89.12     &  36.91  \\
\midrule

Vicuna 7B & 5K G + 15K T       & 88.06    & \textbf{35.59}   \\

Vicuna 7B & 10K G + 10K T              & 89.30    & 34.29  \\

Vicuna 7B & 15K G + 5K  T              & \textbf{89.80}    & 34.01  \\
\bottomrule
\end{tabular}
\caption{The VQA performances of models tuned with different combinations of instructions on PathVQA. G: Generation-based, T: Template-based.}
\label{description}
\end{table}

\begin{table}[!t]
\centering
\begin{tabular}{cccc}
\toprule
      Instruction                            & Recall   & Precision & F1-score  \\
\midrule
15K G + 30K T  & \textbf{54.33}  & 40.74 & 43.56 \\

15K G   & 52.01  & \textbf{43.39} & \textbf{44.21}\\

30K T   & 26.71  & 19.70 & 21.08 \\

\midrule

5K G + 15K T     & 49.28  & 41.00 & 41.71\\

10K G + 10K T     & 52.22  & \textbf{44.35} & 44.92 \\

15K G + 5K T  & \textbf{53.40} &44.17 &\textbf{45.23} \\
\bottomrule
\end{tabular}
\caption{The VQA performances of Vicuna with different combinations of instructions on QUILT-VQA. G: Generation-based, T: Template-based.}
\label{RatioQUILTVQA}
\end{table}

\begin{table}[htbp]
  \centering
  \begin{tabular}{@{}p{0.1\linewidth} p{0.4\linewidth}p{0.4\linewidth}@{}}
    \toprule
    \textbf{Caption:} & Apoptotic keratinocytes are present within the epidermis, consistent with a clinical impression of pityriasis lichenoides at variola formis acuta (PLEVA). PLEVA is characterized by parakeratosis and lymphocytic exocytosis, and may also show lichenoid interface dermatitis.  & \multirow{3}{*}{\includegraphics[width=0.9\linewidth]{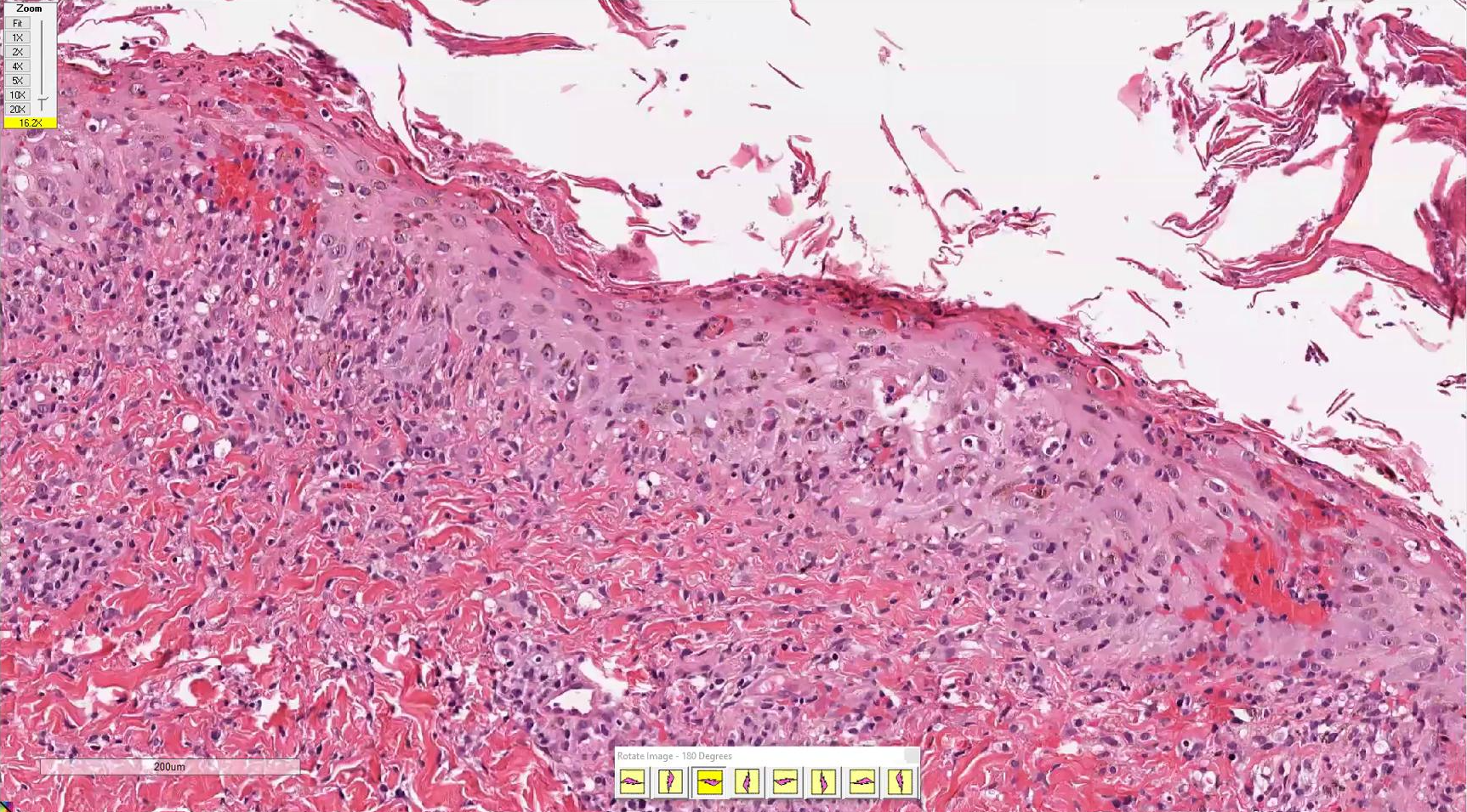}}\\
 
    \midrule

    &\multicolumn{2}{p{0.8\linewidth}}{\textbf{Question and answer generated with PVLM-oriented prompt:}} \\
    \textbf{Question:} & \multicolumn{2}{p{0.8\linewidth}}{What can you tell me about the presence of apoptotic keratinocytes in the image?}\\
   \textbf{Answer:} & \multicolumn{2}{p{0.8\linewidth}}{The image shows the presence of apoptotic keratinocytes within the epidermis. This finding is consistent with a clinical impression of pityriasis lichenoides at variola formis acuta (PLEVA). Apoptotic keratinocytes are cells undergoing programmed cell death, and their presence in the epidermis suggests an abnormality in the cellular turnover process.}\\
    \textbf{Question:} & \multicolumn{2}{p{0.8\linewidth}}{What are the characteristic features of PLEVA?}\\
    \textbf{Answer:} & \multicolumn{2}{p{0.8\linewidth}}{PLEVA is characterized by parakeratosis, which is an abnormal retention of nuclei in the outermost layer of the epidermis, lymphocytic exocytosis, which is the migration of immune cells from blood vessels into the epidermis, and lichenoid interface dermatitis, which is the infiltration of immune cells at the junction between the epidermis and dermis.}\\
    \midrule
    &\multicolumn{2}{p{0.8\linewidth}}{\textbf{Question and answer generated with non-PVLM-oriented prompt (Now you are a pathologist, and your task is to generate question-answer pairs based on the captions of the images.):}}\\
   \textbf{Question:} & \multicolumn{2}{p{0.8\linewidth}}{  What type of cells are present within the epidermis that are consistent with PLEVA?}\\
  \textbf{Answer:} & \multicolumn{2}{p{0.8\linewidth}}{ Apoptotic keratinocytes are present within the epidermis, consistent with PLEVA. }\\
    \textbf{Question:} &  \multicolumn{2}{p{0.8\linewidth}}{What does PLEVA stand for?}\\
    \textbf{Answer:} & \multicolumn{2}{p{0.8\linewidth}}{PLEVA stands for pityriasis lichenoides et variolaformis acuta.}\\

\bottomrule
  \end{tabular}
  \caption{Qualitative comparison of the questions and answers respectively generated with PVLM-oriented prompt and non-PVLM-oriented prompt. (Image source: QUILT-VQA \cite{Quilt-LLaVA})}
    \label{QAcompared}
\end{table}

\begin{table}[!t]
\centering
\begin{tabular}{cccc}
\toprule
LLM       & \begin{tabular}[c]{@{}l@{}}PVLM-oriented \\ prompts\end{tabular} & Closed-end & Open-end \\
\midrule
FlanT5XL  & \ding{55}                     & 85.67     & 12.64    \\
FlanT5XL  & \ding{52}                     & \textbf{86.80}      & \textbf{26.48}    \\
\midrule
Vicuna 7B & \ding{55}                     & 88.59     & 31.79    \\
Vicuna 7B & \ding{52}                     & \textbf{88.76}     & \textbf{35.08}   \\
\bottomrule
\end{tabular}
\caption{VQA performances with and without PVLM-oriented prompts.}
\label{well-designed}
\end{table}

\begin{table}[!t]
\centering
\begin{tabular}{cccc}
\toprule
LLM       & ~Subset~ & ~~Closed-end~~ & ~~Open-end~~ \\
\midrule
Vicuna 7B & 1     & 89.65     & 32.78    \\
Vicuna 7B & 2     & 88.59     & 32.65    \\
Vicuna 7B & 3     & 88.50      & 32.25    \\
\bottomrule
\end{tabular}
\caption{The VQA performances of model trained with different subsets (5K) of the generation-based instruction data.}
\label{groups}
\end{table}

\begin{table}[!t]
    \centering
    \begin{tabularx}{\textwidth}{X}
        \toprule
        Messages = [ \{``role" : ``system",``content": ``As a specialized AI assistant focusing on pathological images, you will receive textual descriptions (caption) of figures. Please note that you do not have access to the actual images. Your task is to generate a set of question-and-answer (QA) pairs between the person inquiring about the images (user) and you as the assistant responding. The QA should be conducted as if both the user and the assistant are examining the images, without referring to textual information.\\
  The following are the requirements for generating question-and-answer pairs:
  
    \begin{itemize}[noitemsep, topsep=0pt]
        \item Avoid referencing dates or magnification ratios.
        \item Focus on visual descriptions, including organizational structure, cellular morphology, potential pathological changes, location, etc.
        \item Avoid using phrases such as ``mention", ``title", ``context", or ``narrator". Instead, refer to information as being ``in the image."
        \item When responding to questions, adopt an objective and responsible attitude, avoiding overconfidence, and refrain from providing medical advice or diagnostic information. Encourage users to consult healthcare professionals for more accurate advice.
    \end{itemize}
      The content should include 4-5 question-and-answer pairs related to visual aspects of the images. " \},\\
    \{few-shot examples\},\\
    \{ ``role": ``user", ``content": description\} ] \\
        \bottomrule
    \end{tabularx}
    \caption{The proposed PVLM-oriented prompt for generating high-quality pathological QA instruction data.}
    \label{prompt}
\end{table}

\begin{table}[htbp]
  \centering
  \begin{tabular}{@{}p{0.1\linewidth} p{0.4\linewidth}p{0.4\linewidth}@{}}
    \toprule
     \textbf{Caption:} & Active inflammation in the stomach can cause epigastric pain and is often associated with H. pylori infection. Successful treatment of H. pylori can lead to regression of the inflammation and healing of the stomach. The described condition is H. pylori gastritis or chronic active gastritis with Helicobacter pylori organisms seen on H and E stain.  & \multirow{3}{*}{\includegraphics[width=0.95\linewidth]{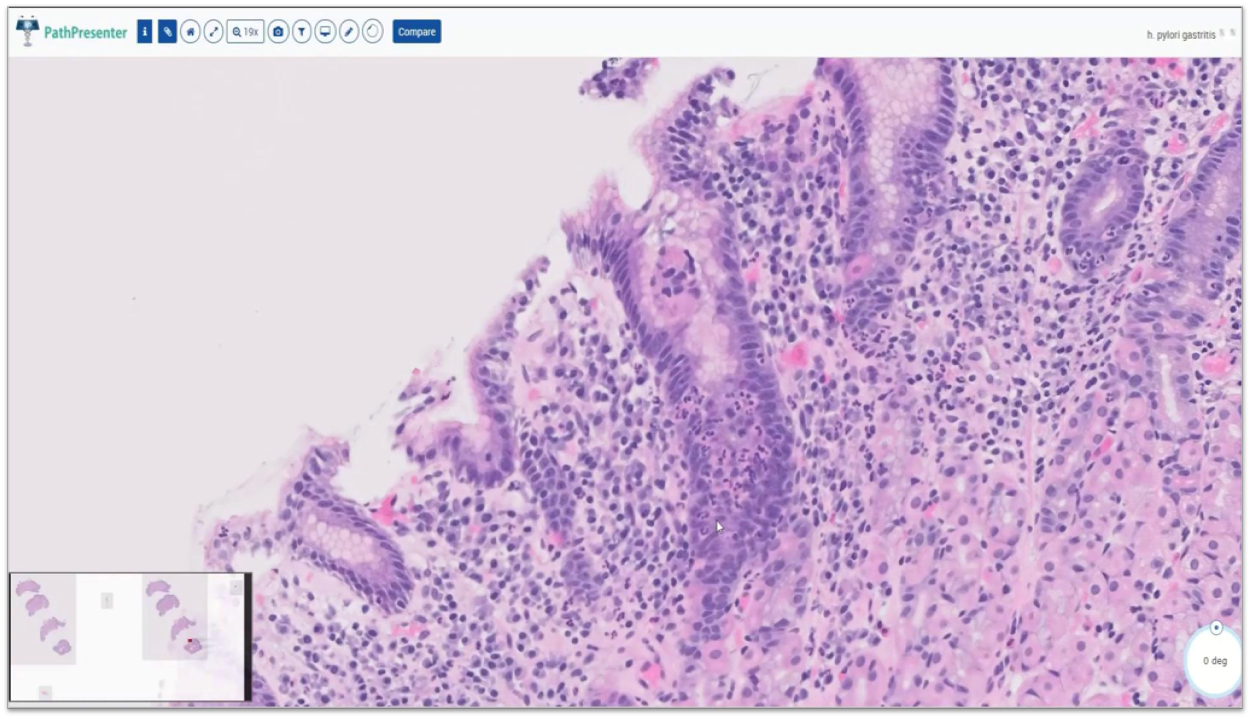}}\\
    \midrule

    \textbf{Question: } & \multicolumn{2}{p{0.8\linewidth}}{What is the described condition?}\\
    \textbf{Answer:}& \multicolumn{2}{p{0.8\linewidth}}{The described condition is H. pylori gastritis or chronic active gastritis. This is an inflammation of the stomach that is often caused by Helicobacter pylori infection.}\\
    \textbf{Question: } & \multicolumn{2}{p{0.8\linewidth}}{What are the symptoms associated with active inflammation in the stomach?}\\
     \textbf{Answer:}&\multicolumn{2}{p{0.8\linewidth}}{Active inflammation in the stomach can cause epigastric pain, which is pain in the upper abdomen. Other symptoms that may be associated with this condition include nausea, vomiting, bloating, and loss of appetite.}\\
     \textbf{Question: } &\multicolumn{2}{p{0.8\linewidth}}{What can lead to regression of the inflammation and healing of the stomach in H. pylori gastritis?}\\
     \textbf{Answer:}&\multicolumn{2}{p{0.8\linewidth}}{Successful treatment of H. pylori infection can lead to regression of the inflammation and healing of the stomach. By eradication of the H. pylori bacteria, the inflammatory response can be reduced, allowing the stomach tissue to heal.}\\
     \textbf{Question: } &\multicolumn{2}{p{0.8\linewidth}}{What staining technique was used to visualize the H. pylori organisms?}\\
     \textbf{Answer:}&\multicolumn{2}{p{0.8\linewidth}}{The H. pylori organisms were visualized using Hematoxylin and Eosin (H\&E) staining technique. This staining method allows for the detection of bacteria and other microscopic structures within the tissue.}\\
\bottomrule
  \end{tabular}
  \caption{An example of the generated question and answer as instructions based on the caption. (Image source: Quilt-1M \cite{Quilt-1M})}
    \label{exampleQA}
\end{table}

\FloatBarrier
\bibliography{Ref}
\end{document}